\pdfoutput=1

\documentclass[10pt,twocolumn,letterpaper]{article}

\usepackage{graphicx}
\usepackage{amsmath}
\usepackage{amssymb}
\usepackage{booktabs}
\usepackage{cite}
\usepackage[accsupp]{axessibility}  

\usepackage[pagebackref,breaklinks,colorlinks]{hyperref}

\usepackage[capitalize]{cleveref}
\crefname{section}{Sec.}{Secs.}
\Crefname{section}{Section}{Sections}
\Crefname{table}{Table}{Tables}
\crefname{table}{Tab.}{Tabs.}

\date{}

\begin{document}

\title{Multi-modal Expression Recognition with Ensemble Method}

\author{
Chuanhe Liu\textsuperscript{1, $\ast$, $\dagger$}, Xinjie Zhang\textsuperscript{2, $\ast$}, Xiaolong Liu\textsuperscript{1}, Tenggan Zhang\textsuperscript{2}\\ 
Liyu Meng\textsuperscript{1}, Yuchen Liu\textsuperscript{2}, Yuanyuan Deng\textsuperscript{1}, Wenqiang Jiang\textsuperscript{1} \\
 \textsuperscript{1} \small{Beijing Seek Truth Data Technology Co.,Ltd.} \\
 \textsuperscript{2} \small{School of Information, Renmin University of China} \\
}
\maketitle


\def\thefootnote{$\ast$}\footnotetext{These authors contributed equally to this work and should be considered co-first authors.}
\def\thefootnote{$\dagger$}\footnotetext{Corresponding Author.}

\maketitle

\begin{abstract}
   This paper presents our submission to the Expression Classification Challenge of the fifth Affective Behavior Analysis in-the-wild (ABAW) Competition. 
   In our method, multimodal feature combinations extracted by several different pre-trained models are applied to capture more effective emotional information. For these combinations of visual and audio modal features, we utilize two temporal encoders to explore the temporal contextual information in the data. In addition, we employ several ensemble strategies for different experimental settings to obtain the most accurate expression recognition results. 
   Our system achieves the average F1 Score of $0.45774$ on the validation set.
\end{abstract}

\section{Introduction}
Affective computing has a wide spectrum of application requirements in human-computer interaction, security, robotics manufacturing, automation, medical, and communications. Actively creating machines that can understand the feelings, emotions, and behaviors of humans would help them interact with humans in a more intimate way and serve effectively\cite{darwin1998expression}. Facial expressions are one of the most powerful, natural, and pervasive signals that humans use to communicate their emotional state and intentions. Machines can analyze human expressions leading to understanding human emotions. The majority of recent research in emotion recognition is based on deep learning, which requires a large quantity of labeled data. Nowadays, there are several datasets, such as Aff-wild\cite{kollias2017recognition, kollias2019deep, zafeiriou2017aff} and Aff-wild2\cite{kollias2022abaw,kollias2021distribution,kollias2021analysing,kollias2021affect,kollias2020analysing,kollias2019expression,kollias2019face,kollias2019deep,zafeiriou2017aff}, which provide us with large-scale data with high-quality labels, which are convenient for training neural networks and increasing the accuracy of expression recognition.

The information of facial expressions can be mainly obtained from the visual modality. Nevertheless, it is well known that audio modality also contains certain emotional information. The information of unimodal modality may be affected by various noises. To obtain a more complete emotional state, information from multiple modalities can be utilized. The multiple modalities' information can supplement and enhance the information to a degree, improving the recognition ability, generalization, and robustness of the model.

Our system for the expression recognition challenge contains several key components. First, the officially provided cropped aligned image data along with the labels are corresponded to, and the data containing labels but no images are complemented. Second, multiple pre-trained feature extractors are employed to extract visual and audio features. Then, we designed multimodal feature combinations and concatenated multiple features into multimodal feature representations. Two different types of temporal encoders, LSTM\cite{sak2014long} and Transformer\cite{vaswani2017attention}, are applied to extract contextual information from the multimodal features. Several techniques are also utilized for optimization. Finally, we adopted several ensemble strategies to ensemble the experimental results for different settings to raise the accuracy of recognition.

\section{Method}

For a given video $X$, it can be separated into tuo parts, the visual data $X^{vis}$ and the audio data $X^{aud}$. The visual data can be stated as a image frames sequence$\{F_{1}, F_{2}, ..., F_{n}\}$, and $n$ denotes the number of image frames in $X$. The goal of the Expression
Classification Challenge is to predict the sentiment label for each frame in the video. 


\subsection{Pre-processing}
Firstly, the officially provided video data is divided into multiple image frames. For each image frame, the face and facial landmarks are recognized by the face detector. The face part is cropped out according to the bounding box to facilitate the extraction of more accurate emotional information later. In order to be consistent with the official labels provided, we use the cropped and aligned face images provided by the competition for the actual processing.

We matched the labels with the images one by one, and found that some of the images corresponding to the labels did not exist in the cropped and aligned set given by the competition, probably because the face images of the corresponding frames were not detected in the video due to the lighting, angle, and other circumstances. For each of these non-existent images, we complement it by finding the nearest frame in the temporal dimension.

\subsection{Multimodal Feature Representation}


During feature extracting phase, we extract multiple type feature in visual and audio modality. 
The model used for extracting visual features including DenseNet-based\cite{iandola2014densenet} model, IResNet100-based\cite{duta2021improved} model, IResNet100-based\cite{iandola2014densenet} facial action unit (FAU) detection model and the MobileNet-based\cite{howard2019searching} model.
The model used for extracting audio features including eGeMAPS\cite{eyben2015geneva}, ComParE 2016\cite{schuller2016interspeech}, VGGish\cite{hershey2017cnn}, Wav2Vec 2.0\cite{baevski2020wav2vec}, ECAPA-TDNN\cite{desplanques2020ecapa} and HuBERT\cite{hsu2021hubert}.

\subsubsection{Visual Features} 
The first type of visual features is extracted by a pre-trained DenseNet model. Specifically, the DenseNet model is pre-trained on the FER+ and the AffectNet datasets. The dimension of the DenseNet-based visual features is 342. And this kind of feature is denoted as densenet.

The second type of visual feature is MAE-based model\cite{he2022masked}. MAE is a self-supervised model that trains label-free data by masking random patches from the input image and reconstructing the missing patches in the pixel space. We used a face dataset of scale 1.2 million, including DFEW\cite{jiang2020dfew}, Emotionet\cite{DBLP:conf/cvpr/Benitez-QuirozS16}, FERV39k\cite{DBLP:conf/cvpr/WangSHLGZGZ22} and so on, to pre-train the MAE encoder. We denotes this kind of feature as mae, and the dimension of it is 768.

The third type of visual feature is IResNet100-based model, and the dimension of IResNet100-based feature is $512$. 
A large-scale facial expression recognition data which consists of FER+\cite{BarsoumICMI2016}, RAF-DB\cite{li2017reliable}\cite{li2019reliable} and AffectNet\cite{mollahosseini2017affectnet} dataset is utilized to pretrain our facial expression recognition IResNet100-based model, which is denoted as ires100.
And a commercial authorized facial action unit detection dataset is used to pretrain the other IResNet100-based model, which is denoted as fau.

The fourth type of visual feature is MobileNet-based model, and the dimension of MobieNet-based feature is $512$.
We utilized AffectNet\cite{mollahosseini2017affectnet} to train a valence-arousal prediction model in the valence-arousal estimation task of AffectNet.



\subsubsection{Audio Features} 

The first type of audio feature is hand-craft features, which consists of eGeMAPS, ComParE 2016 and fbank. eGeMAPS and ComParE 2016 can be extracted using openSmile, and the dimension of these features are 23 and 130. The dimension of fbank is 80. For convenience, we denotes them as egemaps, compare and fbank.

The second type of audio feature is deep features, which consists of Wav2Vec 2.0, ECAPA-TDNN, VGGish and HuBERT. The dimension of Wav2Vec 2.0 feature is $1024$, the dimension of ECAPA-TDNN feature is $512$, the dimension of VGGish feature is $128$ and the dimension of HuBERT is $512$. We denotes them as wav2vec, ecapatdnn, vggish and hubert respectively.

\subsubsection{Multimodal Fusion}
Given the visual features $f^{v}$ and audio features $f^{a}$ corresponding to a frame, they are first concatenated and then fed into a fully-connected layer to produce the multimodal features $f^{m}$. It can be formulated as follows:

\begin{small}
\begin{equation}
    \centering
    f^{m} = W_{f}[f^{v};f^{a}] + b_{f}
\end{equation}
\end{small}
where $W_{f}$ and $b_{f}$ are learnable parameters. 

%


\subsection{Temporal Encoder}

Due to the limitation of GPU memory, we split the videos into segments at first. Given the segment length $l$ and stride $p$, a video with $n$ frames would be split into $[n/p]+1$ segments, where the $i$-th segment contains frames $\{F_{(i-1)*p+1}, ..., F_{(i-1)*p+l}\}$. 
With the multimodal features of the $i$-th segment $f^{m}_{i}$, we employ a temporal encoder to model the temporal context in the video. Specifically, two kinds of structures are utilized as the temporal encoder, including LSTM and Transformer Encoder.

\subsubsection{LSTM-based Temporal Encoder}
Long and short term memory networks (LSTM) are commonly applied to model sequential dependencies of time sequences. In a practical game, we use LSTM to model the temporal relationships in a sequence of frame images from a video. For the $i$-th video segment $s_{i}$, the multimodal features $f^{m}_{i}$ are directly fed into the LSTM. In addition, the last hidden states of the previous segment $s_{i-1}$ are also fed into the LSTM to encode the context between two adjacent segments. It can be formulated as follows:
\begin{small}
\begin{equation}
    \centering
    g_{i}, h_{i} = \text{LSTM}(f^{m}_{i}, h_{i-1})
\end{equation}
\end{small}
where $h_{i}$ denotes the hidden states at the end of $s_{i}$. $h_{0}$ is initialized to be zeros. To ensure that the last frame of $s_{i-1}$ and the first frame of segment $s_{i}$ are consecutive frames, there is no overlap between two adjacent segments when LSTM is used as the temporal encoder. In another word, the stride $p$ is the same as the segment length $l$.

\subsubsection{Transformer-Based Temporal Encoder}

We used a Transformer Encoder to model the temporal feature in the video segment, which can be formulated as:

\begin{small}
\begin{equation}
    \centering
    g_{i} = \text{TRMEncoder}(f^{m}_{i})
\end{equation}
\end{small}

Unlike LSTM, the transformer encoder just models the context in a single segment and ignores the dependencies of frames between segments. 




\begin{table*}[]
\begin{center}
  \caption{The performance of our method on the validation set.}
  \label{tab:overall_performence}
\begin{tabular}{c|c|c|c}
\hline
Model              & Visual Features               & Audio Features                 & F1      \\ \hline
Transformer        & mae,ires100                   & wav2vec,ecapatdnn,hubert       & 0.38362 \\
Transformer        & mae,ires100,densenet          & ecapatdnn,hubert               & 0.39112 \\
Transformer        & mae,ires100,fau,densenet      & ecapatdnn,hubert               & 0.3938  \\ 
lstm               & mae,ires100                   & wav2vec,ecapatdnn,hubert       & 0.37972 \\
lstm               & fau,ires100                   & ecapatdnn,hubert               & 0.38928 \\
lstm               & mae,ires100                   & wav2vec                        & 0.39385 \\
lstm               & densenet,mae,ires100          & wav2vec,ecapatdnn,hubert       & 0.40178 \\
lstm               & densenet,mae,ires100          & ecapatdnn,hubert               & 0.41410\\ \hline
\end{tabular}
\end{center}
\end{table*}

\subsection{Loss Function}
In the training phase, we utilize the RDrop loss which can be formulated as
\begin{small}
    \begin{equation}
    \begin{aligned}
        \centering
        L^{EXPR} = & \frac{1}{2} * (CE(\hat{y}_{1}, y) + CE(\hat{y}_{2}, y)) + \\
        & \alpha * \frac{1}{2} * (KL(\hat{y}_{1}, \hat{y}_{2}) + KL(\hat{y}_{2}, \hat{y}_{1}))
    \end{aligned}
    \end{equation}
    \label{eq:loss}
\end{small}
where $CE$ denotes cross entropy loss, $KL$ denotes Kullback-Leibler divergence loss. $y_{1}$ and $y_{2}$ denotes the first and the second inference prediction logits, $y$ denotes the label.



\section{Experiments}
\subsection{Dataset} 


Expression(Expr) Classification Challenge in fifth ABAW competition is based on Aff-Wild2, a large-scale dataset. Aff-Wild2 is consists of 548 videos and is annotated with 8 expressions (i.e. neutral, anger, disgust, fear, happiness, sadness, surprise and other). 
As for the feature extractors, we used some other dataset for pretraining, which consists of FER+\cite{BarsoumICMI2016}, RAF-DB\cite{li2017reliable, li2019reliable} and AffectNet\cite{mollahosseini2017affectnet}. 
In addition, an authorized commercial FAU dataset is also used for pretraining visual feature extractor, which consists of 7K images in 15 face action unit categories(AU1, AU2, AU4, AU5, AU6, AU7, AU9, AU10, AU11, AU12, AU15, AU17, AU20, AU24, and AU26).
As for the audio feature extractors, we used some different open-source models to extract features. Wav2Vec 2.0\cite{baevski2020wav2vec}, HuBERT\cite{hsu2021hubert} and ECAPA-TDNN\cite{desplanques2020ecapa} are the deep open-source model for extracting audio features.

\subsection{Evaluation Metric}
According to the competition regulations, we use the average F1 score across 8 categories, which can be formulated as 
\begin{small}
\begin{equation}
    \label{eq:metric}
    \centering
    p = \frac{\sum_{i}^{8}F1(\hat{y}_{i}, y_{i})}{8}
\end{equation}
\end{small}
where $F1$ denotes F1 score, $\hat{y}_{i}$ and $y_{i}$ denotes the $i$-th category of prediction and label respectively.

\subsection{Experiment Settings}

First declare that we used Adam\cite{kingma2014adam} optimizer to train models for 25 epochs.
As for the Transformer model, learning rate is 0.0001, the $\alpha$ in equation\ref{eq:loss} is $5$, the affine dimension is $1024$, the number of Transformer encoder layers is $4$, attention heads number is $4$, the dropout ratio in Transformer encoder layer is $0.3$, the sequence length of one segement is $128$, and the hidden size of head layers are \{512, 256\}.

\begin{table}[]
\begin{center}
  \caption{The performance of our method on the 5-fold cross-validation. The First 4 folds are from train set, and the last fold is the original validation set.}
  \label{tab:5fold}
\begin{tabular}{c|c}
\hline
         & F1       \\ \hline
Fold 1   & 0.43697  \\
Fold 2   & 0.38015  \\
Fold 3   & 0.35646  \\
Fold 4   & 0.39170  \\
Fold 5   & 0.38146  \\
Average  & 0.38935  \\ \hline
\end{tabular}
\end{center}
\end{table}

\subsection{Overall Performance on Validation Set}


Table \ref{tab:overall_performence} shows the results of our method on validation set. Among all the results we post in table, we utilized the same training settings as we described in experiment settsings. As the result we post, different feature combinations can lead to different effects.

\subsection{Model Ensemble}

During the model selection phase, we trained some models with different structures, combinations of features and hyper-parameters, and achieve competitive performence. Vote strategy is employed to improve robustness and perfermence of our final prediction, which is described in table \ref{tab:ensemble}. As the results we post, ensemble different architecture and differnet feature combinations can lead to more benefits on validation set.

\begin{table}[]
\begin{center}
  \caption{The results of each single model and the ensemble of them for the expression prediction task on the validation set.}
  \label{tab:ensemble}
\resizebox{\columnwidth}{!}{
\begin{tabular}{c|c|c}
\hline
Model           & Features                                          & F1                \\ \hline
Transformer     & mae,ires100,wav2vec,ecapatdnn,hubert              & 0.38362           \\
Transformer     & ires100,fau,hubert,wav2vec,ecapatdnn              & 0.35119           \\
Transformer     & ires100,fau,densenet,ecapatdnn,hubert             & 0.36087           \\
Transformer     & mae,ires100,densenet,ecapatdnn,hubert             & 0.39380           \\
LSTM            & densenet,mae,ires100,wav2vec,ecapatdnn,hubert     & 0.40178           \\
LSTM            & mae,ires100,wav2vec,ecapatdnn,hubert              & 0.40832           \\
LSTM            & fau,ires100,ecapatdnn,hubert                      & 0.38928           \\
LSTM            & densenet,mae,ires100,ecapatdnn,hubert             & 0.40889           \\
LSTM            & densenet,mae,ires100,ecapatdnn,hubert             & 0.41410           \\ \hline
Ensemble        &                                                   & \textbf{0.45774}  \\ \hline
\end{tabular}
}
\end{center}
\end{table}

\subsection{Ablation Study}

Table \ref{tab:ablation} shows the abalation study perfermence on validation set, we utilized transformer-based model to compare the benefits of different features and feature combinations. Among all the experimental results, we utilized the same experimental setup in the training phase, except for the feature combinations.

\begin{table}[]
\begin{center}
  \caption{Ablation study of features on the validation set.}
  \label{tab:ablation}
\resizebox{\columnwidth}{!}{
\begin{tabular}{c|c|c}
\hline
Visual                      & Audio                     & F1        \\ \hline
fau,densenet                & ecapatdnn                 & 0.35276   \\ 
fau,densenet                & fbank,ecapatdnn           & 0.33887   \\ 
fau,densenet                & ecapatdnn,hubert          & 0.34391   \\ 
fau,ires100,densenet        & ecapatdnn,hubert          & 0.36266   \\
fau,ires100,densenet        & wav2vec,ecapatdnn         & 0.36944   \\
mae,ires100,densenet        & ecapatdnn,hubert          & 0.39380   \\ \hline
\end{tabular}
}
\end{center}
\end{table}

\section{Conclusion}

In this paper, we propose our framework for the Expression Classification Challenge of the fifth Affective Behavior Analysis in-the-wild (ABAW) Competition. Our approach leverages information from multiple modalities in the spatio-temporal dimension. Various temporal encoders are applied to capture the temporal contextual information in the video. In addition, we design multiple high-quality feature combinations to extract more effective emotional information. Our method achieves a performance of $0.45774$ on the validation set.

\bibliographystyle{unsrt}
\bibliography{main}

\end{document}